\documentclass[]{author-kit/fairmeta}

\usepackage{pifont}
\usepackage{amsmath}
\usepackage{amssymb}
\usepackage{mathtools}
\usepackage{amsthm}

\usepackage{enumitem}
\usepackage{algorithm}
\usepackage{algpseudocode}

\usepackage{array}
\usepackage{makecell}
\usepackage{siunitx}
\usepackage{tabularx}
\usepackage{colortbl}

\usepackage{wrapfig}

\usepackage{fvextra}
\usepackage{xurl}

\def\latent{\mathbf{z}}
\def\noise{\boldsymbol{\epsilon}}
\def\control{\mathbf{C}}
\def\chunksize{W}

\def\map{\mathbf{M}}
\def\object{\mathbf{O}}
\def\textprompt{\mathbf{L}}
\def\egoaction{\mathbf{P}}
\def\rigid{\mathbf{S}}

\def\rotation{\mathbf{R}}
\def\translation{\mathbf{t}}
\def\intrinsics{\mathbf{K}}
\def\oimagedescription{\mathbf{I}^{\text{obj}}}
\def\otext{\mathbf{L}^\text{obj}}
\def\obbox{\mathbf{b}}

\def\IR{\mathbb{R}}

\def\cameraNum{N_{\text{cam}}}

\DeclareMathOperator{\stopgradient}{\text{sg}}

\newcolumntype{L}[1]{>{\raggedright\arraybackslash}m{#1}}
\newcolumntype{C}[1]{>{\centering\arraybackslash}m{#1}}

\crefname{section}{Sec.}{Secs.}
\Crefname{section}{Section}{Sections}
\crefname{subsection}{Sec.}{Secs.}
\Crefname{subsection}{Section}{Sections}
\crefname{subsubsection}{Sec.}{Secs.}
\Crefname{subsubsection}{Section}{Sections}
\crefname{figure}{Fig.}{Figs.}
\Crefname{figure}{Figure}{Figures}
\crefname{table}{Tab.}{Tabs.}
\Crefname{table}{Table}{Tables}
\crefname{equation}{Eq.}{Eqs.}
\Crefname{equation}{Equation}{Equations}
\crefname{appendix}{App.}{Apps.}
\Crefname{appendix}{Appendix}{Appendices}

\newcommand{\method}{M$^{\text{4}}$World}
\newcommand{\longtitle}{A Multi-view Multimodal Driving World Model for Interactive Object Manipulation and Minute-long Streaming}
\newcommand{\shorttitle}{\method{}}

\newcommand{\papertitle}{\shorttitle{}: \longtitle{}}

\title{\papertitle}

\renewcommand\author[2][]{\addtolist[#1]{#2}{\authorlist}{\authorformat}{,\quad}}
\author[1\ast]{Ke~Cheng}
\author[2\ast]{Hanqiao~Ye}
\author[1]{Lei~Shi}
\author[1]{Yahui~Liu}
\author[1]{Yunhan~Shen}
\author[3]{Jingtao~Dong}
\author[1]{Zhenke~Wang}
\author[1]{Wenxuan~Ao}
\author[2]{Weixiang~Xu}
\author[1]{Kaining~Huang}
\author[2\dagger]{Shuhan~Shen}
\renewcommand\affiliation[2][]{\addtolist[#1]{#2}{\affiliationlist}{\affiliationformat}{}}
\affiliation[1]{Meituan,\quad}
\affiliation[2]{Institute of Automation, Chinese Academy of Sciences\\}
\affiliation[3]{Beijing Institute of Technology,\quad}
\affiliation[\ast]{Equal contribution,\quad}
\affiliation[\dagger]{Corresponding author}

\abstract{%
  Driving-world generation has emerged as a core capability for scalable autonomous-driving simulation, yet existing methods remain limited in object-level controllability and long-horizon stability.
  We present \textbf{\method{}}, a \textbf{M}ulti-view and \textbf{M}ultimodal generative driving world model that synthesizes future surround-view video streams and synchronized LiDAR scans while supporting interactive object \textbf{M}anipulation and stable \textbf{M}inute-long streaming.
  Fine-grained object manipulation is realized through a flexible conditioning interface that supports explicit control over both the spatial layout and visual appearance of individual objects.
  Stable minute-long streaming, on the other hand, is achieved through a multi-stage training framework that enables online causal generation in only four denoising steps while maintaining coherent world dynamics throughout extended rollouts.
  Building on these components, we introduce an efficient few-clip post-training as well as a suite of visual reference-conditioned generation models, preserving general generation ability while allowing rare-case customization for long-tail controllability.
  To assess controllability beyond realism, we further introduce an automated VLM-based judging pipeline that evaluates scene-level condition adherence, view-wise object controllability, and cross-view object consistency.
  Comprehensive experiments show that \method{} consistently delivers high generation quality, precise controllability, and stable minute-long streaming.
  Together with downstream long-tail augmentation and scene editing, these results demonstrate the potential of \method{} for controllable, scalable driving simulation.
}

\runningtitle{\longtitle{}}
\firstpageleft{\shorttitle{}}
\firstpageright{July 2026}

\hidecomments 

\begin{document}

\maketitle
\thispagestyle{firstpage}


\section{Introduction}

  \begin{figure}[t]
    \centering
    \includegraphics[width=\linewidth]{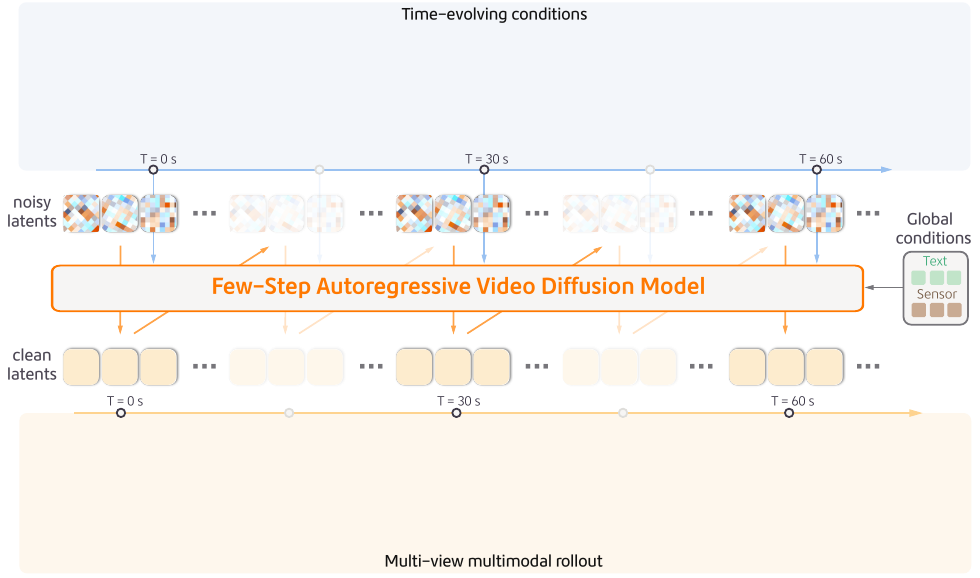}%
    \hspace{-\linewidth}%
    \includegraphics[width=\linewidth]{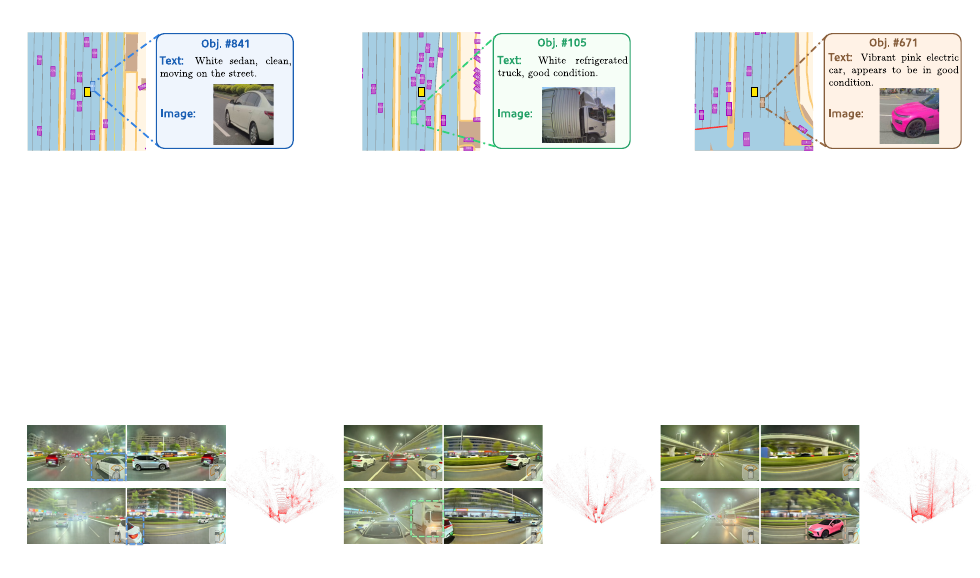}
    \caption{\textbf{\method{} is a few-step autoregressive video diffusion world model for controllable driving simulation.}
      Conditioned on time-evolving control signals as well as global scene and sensor context, it progressively denoises streaming latent chunks and jointly rolls out synchronized multi-view camera videos and LiDAR observations over long horizons, enabling fine-grained object-level control with stable cross-view consistency.
    }
    \label{fig:teaser}
  \end{figure}

  Scaling autonomous-driving testbeds and augmenting training data require controllable simulation environments that overcome the cost and sparsity limitations of real-world data collection.
  Unlike log-replay or reconstruction-only simulators, generative driving world models aim to synthesize action-conditioned, multi-sensor future observations that remain photorealistic, controllable, temporally stable, and responsive to ego-vehicle behaviors over long horizons~\cite{wangDriveDreamerRealworlddrivenWorld2023,gaoMagicDriveV2HighResolutionLong2024,yangGenADGeneralizedPredictive2024,renCosmosDriveDreamsScalableSynthetic2025,yangUniSimNeuralClosedloop2023,zhouXiaomiEVWorld2026,zhengXWorldControllableEgoCentric2026}.
  Such capabilities make them well suited for large-scale policy evaluation, end-to-end training, and targeted augmentation of rare yet safety-critical driving scenarios.

  Reconstruction-only simulators deliver strong geometric fidelity in observed regions but remain fundamentally limited in coverage: they inherit the support of logged sensor trajectories and degrade when extrapolating to unseen viewpoints, novel occlusions, or uncommon scene compositions~\cite{mildenhallNeRFRepresentingScenes2020,kerbl3DGaussianSplatting2023,zhouDrivingGaussianCompositeGaussian2024,yangUniSimNeuralClosedloop2023}.
  Generative and interactive world models have begun to relax this bottleneck by enabling open-ended synthesis and causal rollout~\cite{gaoMagicDriveV2HighResolutionLong2024,aiMAGI1AutoregressiveVideo2025,yinSlowBidirectionalFast2024,zhengXWorldControllableEgoCentric2026}, yet a critical controllability gap remains.
  In current practice, object conditioning is predominantly geometric, relying on 3D bounding boxes or occupancy grids to guide scene composition and object placement, while offering little explicit control over the visual attributes of individual objects~\cite{wenPanaceaPanoramicControllable2023,wangDrivingFutureMultiview2023,gaoVistaGeneralizableDriving2024,zhangHorizonDriveSelfCorrectiveAutoregressive2026}.
  Such attribute-level control is particularly valuable for closed-loop simulation, in which constructing targeted, safety-critical scenarios requires not only placing an object at a desired pose but also specifying precisely what that object should look like.
  Meanwhile, existing long-tail simulation pipelines often confine rare-agent specification to the synthesized initial frame before delegating temporal evolution to an image-to-video model, thereby losing explicit control over pose, orientation, and multi-agent interactions during rollout~\cite{nvidiaNVIDIAOmniDreamsRealTime2026,zhouXiaomiEVWorld2026}.
  Beyond controllability, closed-loop simulation also requires stable long-horizon streaming generation at low latency, while most existing causal adaptation pipelines remain susceptible to train-test exposure bias or high inference latency~\cite{renCosmosDriveDreamsScalableSynthetic2025,gaoMagicDriveV2HighResolutionLong2024}.

  To overcome these limitations, we present \method{}, a generative driving world model that unifies multi-view and multimodal generation with fine-grained object controllability and long-horizon rollout stability.
  At its core, \method{} builds on a shared latent DiT backbone~\cite{peeblesScalableDiffusionModels2023,wanWanOpenAdvanced2025}, in which self-attention captures long-range spatiotemporal dependencies and cross-attention integrates heterogeneous control signals and cross-sensor context into the denoising process.
  This token-based conditioning pathway naturally supports the fine-grained and flexible object-control interface: we extend conventional geometric object conditioning into a more informative object token that fuses 3D box geometry, semantic category, visual appearance description, and textual attributes, allowing control signals to operate at both spatial layout and visual appearance levels.
  To bridge the gap between offline video priors and causal streaming rollout, we develop a progressive training recipe that involves five stages: bidirectional mid-training, teacher-forcing causal adaptation, few-step student ODE initialization, self-forcing with asymmetric DMD, and iterative long-video fine-tuning, complemented by a latent context refresh mechanism for improved inter-chunk consistency.

  Building on these generation capabilities, we further adapt \method{} to practical long-tail simulation, in which safety-critical cases are often specified individually and only a handful of real clips are available.
  We introduce an efficient per-case post-training strategy that combines balanced rare/common sampling with LoRA adaptation~\cite{huLoRALowRankAdaptation2021}, allowing the simulator to bind rare visual and textual attributes to object controls while preserving the base model's general driving-world controllability.
  To accommodate cases where reference observations or edited images are available, we further extend the model into a suite of visual reference-conditioned generation variants, including first-frame conditioning and object completion, so that rare-case scenario construction can be anchored by concrete visual evidence without changing the original control-token interface.

  To make the proposed driving world model scalable in both training and evaluation, we pair it with a dedicated data and evaluation pipeline.
  We curate large-scale, synchronized multi-view camera and LiDAR driving sequences; mine clips spanning challenging weather, motion, and traffic-agent distributions; and automatically annotate scene-level prompts and object-level visual and textual descriptions for fine-grained conditioning.
  Since standard FID/FVD metrics~\cite{heuselGANsTrainedTwo2018,unterthinerAccurateGenerativeModels2019} capture distributional realism but not condition adherence, we develop an automated VLM-based judging pipeline that evaluates scene-level condition adherence, view-wise object controllability, and cross-view object consistency.

  Experiments evaluate \method{} from basic generation to downstream customized simulation for long-tail scenarios.
  On \qty{10}{\second} driving videos, \method{} generates coherent multi-view camera streams and synchronized LiDAR scans, improving FID/FVD over the existing baseline~\cite{gaoMagicDriveV2HighResolutionLong2024} from 41.7/346.1 to 34.8/288.7.
  Under our VLM-based controllability evaluation protocol, \method{} improves object-level visual and textual fidelity from 13.4\% and 11.6\% to 62.7\% and 59.1\%, respectively, demonstrating faithful adherence to object-level visual and textual conditions.
  These fine-grained conditions, in turn, also enhance cross-view consistency during rollout, raising it from 78.9\% to 84.5\%.
  For autoregressive streaming, the causal few-step student sustains \qty{60}{\second} multi-camera rollouts with coherent appearance and dynamics.
  In our throughput benchmark, it achieves 2.3 FPS at $424\times800$ resolution when jointly generating six camera views and one synchronized LiDAR stream on eight A100 GPUs.
  Beyond basic generation, few-clip customization enables our model to synthesize targeted data for a representative long-tail case involving tree-hauling trucks.
  Augmenting \num{50000} real clips with \num{500} synthetic clips improves target recall from 1.0\% to 69.7\% while leaving regular-set mAP essentially unchanged~(66.7\% to 66.8\%).
  Finally, by replacing common objects in the first frame with long-tail counterparts, our visual reference-conditioned generation model propagates the edited appearance across views and throughout the subsequent rollout, enabling zero-shot synthesis of long-tail driving data.

  We summarize our contributions as follows:
  \begin{itemize}
    \item We build a unified driving world model that integrates the core generation capabilities needed for controllable simulation: multi-sensor support, low-latency causal streaming rollout, and a flexible object-level control interface over both spatial layout and visual appearance.
    \item Building on this foundation, we propose two customization pathways for controllable long-tail scenario generation: few-clip post-training for rare cases and visual reference-conditioned generation for scenarios specified through real observations or edited images.
    \item We establish a controllability-oriented evaluation suite with a VLM judge for assessing condition adherence and cross-view consistency.
          Experiments demonstrate substantial gains in generation quality and condition adherence, and further show the effectiveness of \method{} in downstream long-tail perception data augmentation.
  \end{itemize}


\section{Related Work}

  \subsection{World Models for Driving Simulation}
    Closed-loop autonomous-driving simulation primarily hinges on a world model that generates future states of the environment under actor controls.
    We discuss existing world model systems for autonomous-driving simulation along the following two axes.

    \paragraph{Reconstruction-based world models.}
      Reconstruction-based world models use captured multi-view observations to recover accurate driving scene structure and deliver photorealistic renderings.
      In this line of work, NeRF~\cite{mildenhallNeRFRepresentingScenes2020,mullerInstantNeuralGraphics2022,tancikBlockNeRFScalableLarge2022,chengUCNeRFNeuralRadiance2023,irshadNeO360Neural2023,guoStreetSurfExtendingMultiView2023,tonderskiNeuRADNeuralRendering2024} and 3D Gaussian Splatting~\cite{kerbl3DGaussianSplatting2023,wu3DGUTEnablingDistorted2025,weiParkGaussianSurroundview3D2026,liuCityGaussianRealtimeHighquality2024,liHOGaussianHybridOptimization2024} have been the two dominant neural representations.
      Building on these representations, substantial efforts have focused on modeling dynamic 4D driving scenes.
      Some approaches model a dynamic scene by using a deformation network to map time-dependent observations to a canonical space~\cite{yangDeformable3DGaussians2023,wu4DGaussianSplatting2024} or by feeding timestamps as additional inputs to the neural representation~\cite{huang$textitS^3$GaussianSelfSupervisedStreet2024,yangEmerNeRFEmergentSpatialTemporal2023,turkiSUDSScalableUrban2023}.
      To support individual control of dynamic agents, the scene can be factorized into a static background model together with separate moving object models~\cite{zhouDrivingGaussianCompositeGaussian2024,tonderskiNeuRADNeuralRendering2024,chenOmniReOmniUrban2024,yanStreetGaussiansModeling2024,yuanUniGaussiansUnifyingCamera2025,yangUniSimNeuralClosedloop2023,turkiSimULiRealTimeLiDAR2026}.
      To overcome the inefficiency of per-scene optimization, feed-forward frameworks have been proposed to learn generalizable priors across scenes, inferring 3D Gaussian representations in a single forward pass~\cite{yangSTORMSpatioTemporalReconstruction2024,tanUFOUnifyingFeedForward2026,chenDGGTFeedforward4D2025}.
      Despite these advancements, reconstruction-based world models are still constrained by the coverage of sensor data and often struggle to render unobserved regions and generalize to uncommon scene conditions.

    \paragraph{Video generation for autonomous driving.}
      Diffusion-based video generation models~\cite{peeblesScalableDiffusionModels2023,polyakMovieGenCast2025,wanWanOpenAdvanced2025,kongHunyuanVideoSystematicFramework2025,hongCogVideoLargescalePretraining2022,nvidiaCosmosWorldFoundation2025,zhengOpenSoraDemocratizingEfficient2024} unlock the potential of synthesizing diverse driving scenarios beyond the coverage of captured data.
      Motivated by this complementary strength, some methods leverage diffusion priors to synthesize future observations~\cite{gaoMagicDrive3DControllable3D2025,zhaoDriveDreamer4DWorldModels2024} or restore ghost artifacts in novel-view renderings for reconstruction augmentation~\cite{tanExtraGSGeometricAwareTrajectory2025,yangNeoVerseEnhancing4D2026,wuDifix3DImproving3D2025,niReconDreamerCraftingWorld2024}.
      In parallel, a growing body of work seeks to directly generate realistic and controllable driving videos from structured scene representations and driving behaviors.
      These methods have evolved along three main directions.
      The first introduces structured conditions over scene layouts, weather, agents, and ego trajectories to enable realistic and controllable generation~\cite{huGAIA1GenerativeWorld2023,russellGAIA2ControllableMultiView2025,wangDriveDreamerRealworlddrivenWorld2023,gaoVistaGeneralizableDriving2024}.
      The second pursues spatial controllability and multi-view consistency through explicit 3D geometry, BEV-style controls, or panoramic generation~\cite{gaoMagicDriveStreetView2024,gaoMagicDriveV2HighResolutionLong2024,wangDrivingFutureMultiview2023,wenPanaceaPanoramicControllable2023}.
      The third scales toward large-scale generalist predictors and foundation world models that expand training scale, downstream utility, and controllable rollout across camera and LiDAR observations~\cite{yangGenADGeneralizedPredictive2024,renCosmosDriveDreamsScalableSynthetic2025}.
      Taken together, these directions set the stage for a broader shift from open-loop video synthesis to interactive, action-conditioned simulation for autonomous driving.
      Recent systems already operationalize this shift by integrating reconstruction and generation for geometric fidelity~\cite{zhouXiaomiEVWorld2026}, scaling real-time closed-loop generative simulation~\cite{nvidiaNVIDIAOmniDreamsRealTime2026}, improving long-horizon autoregressive stability via self-corrective distillation~\cite{zhangHorizonDriveSelfCorrectiveAutoregressive2026}, and enabling controllable ego-centric multi-camera rollouts for scalable end-to-end evaluation~\cite{zhengXWorldControllableEgoCentric2026}.
      Building on this line of work, our method emphasizes object-centric control to improve both controllability and scenario diversity in generated rollouts.
      Both capabilities are crucial for driving simulation, in which scalable synthesis of long-tail scenarios remains a central challenge.

  \subsection{Real-Time Interactive Video World Models}
    Recent foundation models for offline video generation provide strong generative priors for visual appearance, motion, and scene evolution~\cite{brooksVideoGenerationModels2024,hongCogVideoLargescalePretraining2022,wanWanOpenAdvanced2025,kongHunyuanVideoSystematicFramework2025,zhengOpenSoraDemocratizingEfficient2024}.
    Interactive world models can build on these priors to realize low-latency causal rollouts under user controls such as camera trajectories.
    To this end, recent work combines diffusion modeling with autoregressive~(AR) prediction to adapt video generators for causal, streaming rollout.
    MAGI-1~\cite{aiMAGI1AutoregressiveVideo2025} performs chunk-wise autoregressive generation with progressive denoising.
    CausVid~\cite{yinSlowBidirectionalFast2024} converts a pretrained bidirectional diffusion transformer~\cite{hongCogVideoLargescalePretraining2022} into a few-step causal generator with asymmetric distillation~\cite{yinOnestepDiffusionDistribution2023} and ODE initialization.
    Building on this causal formulation, diverse training strategies have been proposed to reduce the training-inference gap and improve rollout quality and temporal consistency~\cite{cuiSelfforcingMinutescaleHighquality2025,huangSelfForcingBridging2025a,fengOneForcingStableOneStep2026,zhuCausalForcingAutoregressive2026,zhaoMinWMFullStackOpenSource2026}.
    To sustain longer rollouts, Rolling Forcing~\cite{liuRollingForcingAutoregressive2025} expands the diffusion window, whereas LongLive~\cite{yangLongLiveRealtimeInteractive2025} refreshes the KV cache to preserve visual continuity and prompt adherence across scene transitions.
    Meanwhile, attention sink behavior~\cite{xiaoEfficientStreamingLanguage2024,yiDeepForcingTrainingFree2025} has been explored to improve long-range temporal consistency.
    Following prior interactive world models, we adopt a DiT backbone for multi-view, multimodal video generation.
    Fine-grained controllability is achieved via cross-attention, which mediates information exchange between control signals and cross-sensor observations.


\section{Overview}

  \subsection{Problem Formulation}

    We aim to develop a low-latency, long-horizon video world model for autonomous driving that maintains strong spatial and temporal consistency while supporting multi-view and multimodal generation under diverse control signals.
    Our model operates autoregressively in the latent space of a video VAE\@.
    Let $\latent_{1:T}$ denote the latent context and $\control_{T+1:T+\chunksize}$ the control signals for the subsequent $\chunksize$ frames to be generated.
    At each autoregressive step, the model samples the next latent chunk according to
    \begin{equation}
      \hat{\latent}_{T+1:T+\chunksize}
      \sim p_{\theta}\!\left(
      \latent_{T+1:T+\chunksize}
      \mid \latent_{1:T},
      \control_{T+1:T+\chunksize}
      \right),
    \end{equation}
    and appends it to the context for the next step.
    The resulting latent chunk is decoded into temporally aligned surround-view videos and synchronized LiDAR scans represented as range maps.
    We instantiate this framework by adapting an open-source bidirectional text-to-video generation model~\cite{wanWanOpenAdvanced2025} into an autoregressive, controllable, multimodal world model through a multi-stage training strategy.
    The model architecture and training procedure are detailed in \cref{sec:architecture,sec:training}, respectively.

    Unlike most prior work~\cite{gaoMagicDriveStreetView2024,gaoMagicDriveV2HighResolutionLong2024,wangDriveDreamerRealworlddrivenWorld2023}, whose object-level controls remain largely geometric, we introduce fine-grained conditions that complement conventional geometric controls.
    These conditions enable precise control over individual objects and, crucially, controllable generation of rare, long-tail objects that are difficult to capture at scale in real-world driving data~(\cref{sec:longtail}).
    To evaluate this capability at scale, we further introduce a dedicated metric suite for automatically measuring the fidelity of fine-grained control in generated driving scenes~(\cref{sec:metric}).

  \subsection{Model Architecture of \method{}}
    \label{sec:architecture}

\begin{figure}[t]
      \centering
      \includegraphics{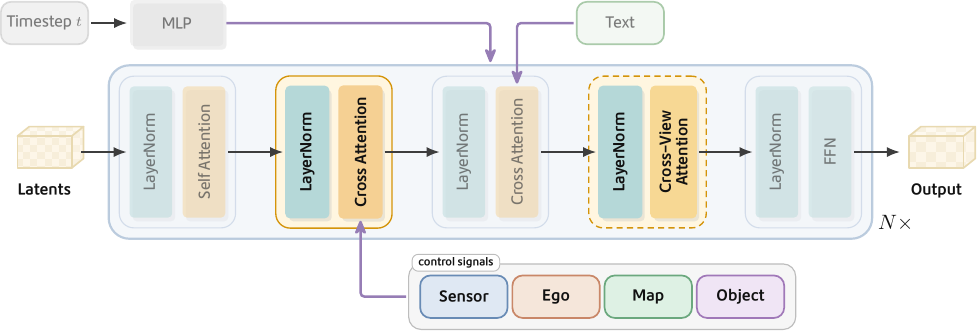}
      \caption{
        \textbf{Architecture overview.}
        The \method{} model adopts a shared DiT backbone with two cross-attention pathways to integrate control signals and cross-sensor context, enabling unified multi-view and multimodal generation under heterogeneous driving conditions.
      }
      \label{fig:transformer_block}
    \end{figure}

    A Diffusion Transformer (DiT)~\cite{peeblesScalableDiffusionModels2023} is employed as the backbone of our latent video world model.
    DiT first patchifies and flattens the video latents into spatiotemporal tokens, processes them through stacked Transformer blocks, and then unpatchifies the outputs back into the latent grid.
    As illustrated in \cref{fig:transformer_block}, within each block, self-attention captures long-range spatiotemporal dependencies, while cross-attention injects heterogeneous control signals into the shared backbone.
    Together, these properties provide more flexible multimodal conditioning and more favorable scaling capacity than U-Net backbones or ControlNet-style auxiliary branches.

    \paragraph{Control signals.}
      We adopt the control signals defined in~\cite{gaoMagicDriveStreetView2024,gaoMagicDriveV2HighResolutionLong2024}, with the per-frame condition set given by $\control_{t}=\{\textprompt, \rigid, \egoaction_{t}, \map_{t}, \object_{t}\}$.
      These signals fall into two categories: global conditions~(temporally consistent) and time-evolving conditions.
      The temporally consistent conditions comprise the scene text prompt $\textprompt$ and the sensor parameters $\rigid$, both fixed throughout a generated sequence.
      The camera-rig condition is defined as $\rigid_{\text{cam}}=\{(\intrinsics_i,\rotation_i,\translation_i)\}_{i=1}^{\cameraNum}$, where $\intrinsics_i$ is the intrinsic matrix of camera $i$, and $(\rotation_i,\translation_i)$ specifies its pose relative to the LiDAR coordinate frame.
      The time-evolving conditions include the ego-vehicle poses $\egoaction_{t}=\{(\rotation_t, \translation_t)\}$, the static BEV map $\map_{t}\in \{0, 1\}^{w\times h\times c}$ representing a $w\times h$-meter traffic area with $c$ classes, and object attributes $\object_{t}$, which evolve over time to control the scene dynamics in each frame.

      In particular, at timestamp $t$, $\object_{t}=\{\mathbf{o}_{t,n}\}_{n=1}^{N_t}$ collects the attributes of all $N_t$ objects in the scene.
      At frame $t$, each object representation $\mathbf{o}_{t,n}$ includes not only the conventional 3D bounding box $\obbox_{t,n} = \{(x_j, y_j, z_j)\in\IR^3\}_{j=1}^8$ and semantic category $c$, but also an image description $\oimagedescription_{t,n}$ and a fine-grained textual description $\otext_{t,n}$, enabling precise appearance-level control beyond object position and orientation.
      \begin{itemize}
        \item \textbf{Text tokens}: We employ the umT5 model~\cite{chungUniMaxFairerMore2023} to encode the high-level scene description $\textprompt$.
        \item \textbf{Sensor tokens} and \textbf{ego tokens}:
              Both the camera-rig parameters $\rigid_\text{cam}$ and ego poses $\egoaction_{t}$ can be represented as sequences of 3D vectors.
              Each camera condition is formed as $[\intrinsics_i;\rotation_i;\translation_i^{\top}]\in\IR^{7\times3}$, whereas the ego pose at time $t$ is defined relative to the first frame as $[\rotation_t;\translation_t^{\top}]\in\IR^{4\times3}$.
              We apply Fourier embedding~\cite{mildenhallNeRFRepresentingScenes2020} to each 3D vector and use separate MLPs to obtain control tokens for the camera parameters and ego poses.
              The LiDAR sensor $\rigid_\text{lidar}$ serves as the reference coordinate frame and therefore does not require an explicit camera-like geometric parameterization.
              We instead represent it using a randomly initialized learnable token.
        \item \textbf{Map tokens}: Unlike prior methods that inject 2D structural conditions through ControlNet-style auxiliary branches, which often introduce substantial parameter and computational overhead, we adopt a lightweight tokenizer that patchifies and flattens the 2D BEV map $\map_{t}$ into a sequence of control tokens.
        \item \textbf{Object tokens}:
              The control token for each object $\mathbf{o}_{t,n}$ combines the four complementary types of information described above: a 3D bounding box, a class label, an image description, and a textual description.
              For the class label, following prior practice~\cite{liGLIGENOpenSetGrounded2023}, we use the pooled embedding of its class name as the label embedding.
              For the 3D box $\obbox_{t,n}\in\IR^{8\times3}$, represented by the coordinates of its eight corner points, we apply Fourier embedding~\cite{mildenhallNeRFRepresentingScenes2020} to each point and encode the resulting features with an MLP\@.
              For the image description $\oimagedescription_{t,n}$, we extract an appearance feature using SigLIP-V2~\cite{tschannenSigLIP2Multilingual2025}.
              For the textual description $\otext_{t,n}$, we obtain its embedding with the umT5 model~\cite{chungUniMaxFairerMore2023}.
              Finally, we concatenate these embeddings and compress them with an MLP, producing a control token for each individual object that captures both its geometric layout and fine-grained appearance and semantics.
      \end{itemize}

    \paragraph{LiDAR modality support.}
      We project each LiDAR scan into a range map, a 2D image whose pixels encode the radial distance to the nearest surface along each ray.
      This representation allows us to directly reuse the DiT backbone for parallel multimodal generation.
      The raw range map has resolution $128\times 1300$, where the vertical dimension corresponds to 128 LiDAR beams and the horizontal dimension corresponds to azimuth samples.
      Before VAE encoding, we normalize LiDAR distances within the valid range of $[0,100]$ meters to $[-1,1]$ and resize the range map to the same spatial resolution as the image input.
      We then reuse the same video VAE to encode the range map into a latent representation, aligning LiDAR and video observations in a unified latent space.

    \paragraph{Multi-condition and multi-view aggregation.}
      Building on a generic text-to-video DiT architecture, we retain its original text-conditioning pathway to integrate the high-level scene description $\textprompt$.
      The driving-specific control tokens derived from sensor parameters $\rigid$, ego poses $\egoaction_t$, BEV maps $\map_t$, and object attributes $\object_t$ are then injected into the video backbone through an additional cross-attention layer in each DiT block.
      To further enhance cross-view consistency in both surround-view and LiDAR generation, we follow MagicDriveV2~\cite{gaoMagicDriveStreetView2024} and introduce another cross-attention layer for view-wise feature aggregation.
      Specifically, for the latent tokens of each target view, this layer attends to the latents from all views with overlapping fields of view, allowing each view to incorporate shared observational context while preserving its own view-specific structure.
      In our implementation, we insert this cross-view attention every five Transformer layers as a design trade-off between effectiveness and efficiency.


\section{Long-Horizon Streaming Rollout}
  \label{sec:training}

  We initialize our model from the bidirectional Wan2.1-T2V model~\cite{wanWanOpenAdvanced2025}, which provides strong visual generation priors but is originally designed for offline text-to-video synthesis.
  Transforming it into a low-latency interactive driving world model that supports controllable multimodal generation, causal streaming rollout, and stable long-horizon extrapolation requires several dedicated training stages.

  \subsection{Bidirectional Mid-Training for Driving Scene Dynamics}
    \label{sec:bidirectional}
    The first stage transfers the base video generator to the driving domain by training it on autonomous-driving data with the full set of control signals introduced in \cref{sec:architecture}.
    Beyond domain adaptation, this mid-training stage expands the model's input-output interface: it learns to respond to multiple driving controls while generating synchronized multi-view and multimodal observations.
    We retain the original bidirectional attention during this stage, allowing the model to exploit both past and future context within each training clip and preserve its ability to model driving-scene dynamics before causal adaptation.

    Specifically, we optimize the model with a rectified-flow objective~\cite{liuFlowStraightFast2022}.
    Let $\mathbf{x}_0$ denote a clean latent training sample and $\noise\sim\mathcal{N}(\mathbf{0},\mathbf{I})$ denote Gaussian noise with the same shape.
    For a randomly sampled interpolation time $t\sim\mathcal{U}(0,1)$, we construct an intermediate latent state by linearly mixing the two endpoints:
    \begin{equation}
      \mathbf{x}_t = (1-t)\noise + t\mathbf{x}_0.
    \end{equation}
    Under this straight transport path, the target velocity is constant along the trajectory and points from the noise sample to the data sample, i.e.,
    \begin{equation}
      \mathbf{v}^{*} = \mathbf{x}_0 - \noise.
    \end{equation}
    The DiT backbone is trained to predict this velocity field conditioned on the driving control signals $\control$, yielding the objective
    \begin{equation}
      \mathcal{L}_{\mathrm{rf}}(\theta)
      =
      \mathbb{E}_{\mathbf{x}_0,\noise,t}
      \left[
        \left\|
        \mathbf{v}_{\theta}(\mathbf{x}_t,t; \control) - (\mathbf{x}_0-\noise)
        \right\|_2^2
        \right].
    \end{equation}
    During training, we begin with visual-only generation to establish a stable driving-scene prior.
    We mix video clips at three spatial resolutions, $224\times400$, $424\times800$, and $576\times1024$, with sequence lengths ranging from 13 to 93 frames.
    Similarly, we randomly sample the camera configuration at each training iteration, using up to 10 views.
    We then enable LiDAR range-map generation after this training has sufficiently progressed.

    At generation time, samples are obtained by starting from Gaussian noise and numerically following the learned velocity field toward the data distribution.
    A practical issue is that standard classifier-free guidance~(CFG) often causes over-saturation in generated images; for LiDAR range maps, this effect translates into systematic depth shifts that are unacceptable for driving simulation.
    We therefore adopt Adaptive Projected Guidance~(APG)~\cite{sadatEliminatingOversaturationArtifacts2025} for LiDAR sampling, which introduces no additional computation over standard guidance while effectively improving the quality of generated range maps.

  \subsection{Causal Student Distillation for Autoregressive Rollout}
    \label{sec:causal-distillation}

    \paragraph{Teacher Forcing~(TF).}
      After bidirectional mid-training, \emph{Teacher Forcing} adapts the model from offline generation to autoregressive streaming rollout by applying a causal mask to temporal attention.
      Specifically, given a training sequence, we sample a \emph{ground-truth} video latent $\latent_{1:T}$ as the context prefix and use the subsequent chunk $\latent_{T+1:T+\chunksize}$ as the prediction target.
      During this stage, the model is trained with causal attention, so the target chunk can only condition on the observed prefix and the corresponding future control signals $\control_{T+1:T+\chunksize}$.
      Following the same rectified-flow parameterization, we perturb the target chunk with Gaussian noise $\boldsymbol{\epsilon}\sim\mathcal{N}(\mathbf{0},\mathbf{I})$ at time $t\sim\mathcal{U}(0,1)$:
      \begin{equation}
        \mathbf{x}_t = (1-t)\boldsymbol{\epsilon} + t\latent_{T+1:T+\chunksize}.
      \end{equation}
      The model predicts the velocity from the noisy chunk to the clean target while being conditioned on both driving signals $\control_{T+1:T+\chunksize}$ and the teacher-forced history $\latent_{1:T}$:
      \begin{equation}
        \mathcal{L}_{\mathrm{tf}}(\theta)
        =
        \mathbb{E}_{\latent_{1:T+\chunksize},\noise,t}
        \left[
          \left\|
          \mathbf{v}_{\theta}
          \left(
          \mathbf{x}_t, t;
          \control_{T+1:T+\chunksize}, \latent_{1:T}
          \right)
          -
          \left(\latent_{T+1:T+\chunksize}-\boldsymbol{\epsilon}\right)
          \right\|_2^2
          \right].
      \end{equation}
      By always providing the clean historical context during training, \emph{Teacher Forcing} gives the model a stable supervised signal for few-step chunk prediction before it is exposed to its own generated histories in long-horizon rollout.

    \paragraph{Causal ODE initialization.}
      Following prior work~\cite{yinSlowBidirectionalFast2024,zhaoMinWMFullStackOpenSource2026,zhuCausalForcingAutoregressive2026}, we use the AR diffusion model trained with \emph{Teacher Forcing} to supervise a four-step causal student model.
      This initialization reduces rollout latency and provides a stronger starting point for the subsequent \emph{Self-Forcing} and asymmetric distribution matching distillation~(DMD) stage.
      Concretely, the teacher model first produces offline ODE trajectories for target chunks under the ground-truth history $\latent_{1:T}$ and future controls $\control_{T+1:T+\chunksize}$.
      We then sample an intermediate noisy latent $\mathbf{x}_{t}$ along these trajectories over a subset of $t$ values and train the student generator $G_{\theta}$ to map it back to the clean target chunk $\latent_{T+1:T+\chunksize}$:
      \begin{equation}
        \mathcal{L}_{\mathrm{ode}}(\theta)
        =
        \mathbb{E}_{\latent_{1:T+W},t,\mathbf{x}_t}
        \left[
          \left\|
          G_{\theta}
          \left(
          \mathbf{x}_{t}, t;
          \control_{T+1:T+\chunksize}, \latent_{1:T}
          \right)
          -
          \latent_{T+1:T+\chunksize}
          \right\|_2^2
          \right].
      \end{equation}

    \paragraph{Self-Forcing and asymmetric DMD.}
      To close the train-test distribution gap, the four-step causal student model is further trained with \emph{Self-Forcing}~\cite{huangSelfForcingBridging2025a}, where the model is exposed to its own generated history during training.
      As a form of holistic video-level supervision, we employ DMD~\cite{yinOnestepDiffusionDistribution2023,yinImprovedDistributionMatching2024} to align the distribution of the causal student's output with that of the bidirectional teacher.
      The objective is formulated as the reverse KL divergence, whose gradient can be approximated by the difference between the teacher's and student's score functions:
      \begin{equation}
        \mathcal{L}_{\mathrm{dmd}}(\theta) = \mathbb{E}_{\hat{\latent}, t, \hat{\latent}_t}\left[
          \frac{1}{2}\left\|
          \hat{\latent} - \stopgradient\left[\hat{\latent} -
            \left(s_{\text{fake}}(\hat{\latent}_t, t) - s_{\text{real}}(\hat{\latent}_t, t)\right)
            \right]
          \right\|_2^2
          \right],
      \end{equation}
      where $\stopgradient[\cdot]$ denotes the stop-gradient operator.
      The full video sequence $\hat{\latent}$ is generated by the student $G_{\theta}$ through self-rollout and then perturbed into $\hat{\latent}_t$ through the forward diffusion process.
      The score of $\hat{\latent}_t$ in the real data distribution is estimated by the frozen teacher model, while the score of $\hat{\latent}_t$ in the student distribution is estimated by a fake score network~(critic model) trained online.
      Since the critic provides the student-distribution score $s_{\text{fake}}$, inaccurate critic estimates can directly produce unstable or biased DMD gradients.
      Therefore, to keep the critic closely aligned with the evolving distribution of the student generator, we follow DMD2~\cite{yinImprovedDistributionMatching2024} and update the critic five times for every student-generator update.
      In addition, with a probability of 10\% during training, we combine the DMD objective with the supervised denoising loss $\mathcal{L}_{\mathrm{denoise}}$ on ground-truth video latents, which empirically improves training stability and mitigates mode collapse.

    \paragraph{Latent context refresh.}
      Although rolling the KV cache within a fixed-size window~\cite{huangSelfForcingBridging2025a} enables efficient extrapolation through context reuse, salient flickering artifacts still appear at chunk boundaries due to the distribution mismatch.
      Inspired by image-to-video conditioning and existing KV-recache implementations for overlapping frames between consecutive sliding windows~\cite{yinSlowBidirectionalFast2024,aiMAGI1AutoregressiveVideo2025}, we introduce a simple latent context refresh mechanism at inference time.
      Specifically, after denoising each chunk, we feed the latent representation of its final frame back into the network as additional contextual input for the next chunk.
      This differs from pure autoregressive KV caching, where cached keys and values are computed only from historical tokens and are reused unchanged.
      In our setting, the refreshed contextual input participates in attention with the noisy latent tokens of the next chunk; its keys and values therefore depend on the current chunk noise and must be recomputed rather than simply rolled forward.
      By allowing the next chunk to attend to an explicitly refreshed visual anchor, this mechanism empirically improves inter-chunk visual consistency in a training-free manner.

  \subsection{Iterative Fine-Tuning on Long Videos}
    \label{sec:long-tuning}
    Although the causal student can autoregressively roll out beyond the short clips seen during training, the previous DMD stages still operate on clips of at most 93 frames.
    During longer rollouts, the model repeatedly conditions on its own predictions through a finite temporal window, so accumulated errors can gradually corrupt the self-generated context and degrade generation quality.
    Following LongLive~\cite{yangLongLiveRealtimeInteractive2025}, we further adapt the student with iterative long-video fine-tuning, using longer self-rollouts and local temporal supervision to better align the train-time context distribution with that encountered during autoregressive inference.
    Specifically, the student generates a 600-frame self-rollout, during which we iteratively fine-tune it on each newly generated short clip while treating the previously generated frames as causal context.

    Moreover, since the control signals include time-evolving conditions such as ego motion, maps, and object states, each rollout step is trained to incorporate control tokens for the current timestamp.
    Newly injected controls can therefore take effect naturally during autoregressive generation, without requiring an explicit KV-recache operation as in~\cite{yangLongLiveRealtimeInteractive2025}.

  \subsection{Infrastructure Optimizations}
    Having described how the multi-stage training recipe enables low-latency streaming rollout with a four-step causal student, we further introduce infrastructure-level optimizations used during training and inference.

    \paragraph{Denoising DiT.}
      We optimize the denoising of long spatiotemporal token sequences using balanced sequence parallelism as in~\cite{chenLongLive20NVFP4Parallel2026}.
      This strategy distributes tokens along the sequence dimension across multiple GPUs to reduce per-GPU activation memory while preserving full-sequence computation.
      Moreover, since our model delivers multi-view and multimodal generation, we introduce \emph{sensor parallelism}, which distributes full sequences of different sensors across GPUs for parallel computation.
      Unlike sequence parallelism, sensor parallelism requires communication only at cross-view attention layers.
      In practice, we find that inserting one such layer every five Transformer blocks is sufficient; we therefore prioritize sensor parallelism because it requires less frequent communication.

    \paragraph{VAE decoding.}
      For final VAE decoding, another inference bottleneck, we adopt the asynchronous pipeline of LongLive2.0~\cite{chenLongLive20NVFP4Parallel2026} and parallelize decoding across sensors and modalities.

\section{Efficient Post-Training for Long-Tail Scenarios}
  \label{sec:longtail}
  Although the model can learn associations between geometric controls and scene- or object-level prompts from abundant common examples, many safety-critical objects and other objects of interest in driving simulation are intrinsically long-tailed.
  In our data, a rare prompted context often appears only a few times, and in many cases, fewer than five clips contain the target object.
  We find that training on general driving datasets alone does not guarantee reliable controllability for long-tail objects.

  To address this limitation, we perform supervised fine-tuning of our base model separately for each rare case.
  For each rare case, we construct a fine-tuning set with a balanced sampling strategy: 50\% from the target rare-object clips and the remaining 50\% from ordinary driving clips.
  Upsampling the target rare-object clips provides a direct learning signal that binds the rare object's visual and textual prompts to the corresponding object token, thereby strengthening the model's ability to render the target object within the specified 3D box.
  At the same time, mixing common training data prevents the adaptation from drifting away from the general driving distribution.

  Full-parameter fine-tuning on only a few rare clips can still deteriorate output diversity and weaken the pretrained model's broad controllability.
  We therefore employ LoRA~\cite{huLoRALowRankAdaptation2021} adapters for this per-case adaptation while keeping the base model frozen.
  This parameter-efficient update requires only a few hundred iterations to learn the rare-object attributes while preserving the base model's original control over object position, weather, and illumination.

\section{Visual Reference-Conditioned Generation}
  \label{sec:visual_conditioned_suite}

  The condition set $\control$ used by the preceding model contains only the basic control signals required for controllable driving-scene generation; it does not include scene-level visual appearance information.
  In practical applications, however, additional scene-level visual observations are sometimes available and should be injected as appearance constraints for subsequent generation.

  We therefore extend the bidirectional teacher trained in \cref{sec:training} into a suite of visual reference-conditioned generation models.
  There are three variants corresponding to different forms of available visual information:

  \begin{itemize}
    \item \textbf{First-frame multi-view conditioning}: Given the first frame from all views, the model predicts subsequent multi-view frames.
    \item \textbf{First-frame single-view conditioning}: Given the first frame from a single view, the model synthesizes synchronized multi-camera views and predicts future frames.
    \item \textbf{Object completion}: Given a complete video with one object masked out, the model completes the missing object according to the control signals.
  \end{itemize}

  All three variants share the same conditioning mechanism.
  The video VAE encodes the provided scene-level visual information into a reference latent, which is spatially and temporally expanded to match the resolution and length of the initial noise latent.
  We concatenate this reference latent with the latent noise along the channel dimension before feeding it into the DiT backbone.
  For each task, we further construct a task-specific binary mask to indicate the observed visual region or the target completion region and inject it through the same channel-wise concatenation.
  This design keeps the original control-token pathway unchanged while allowing the generation process to be anchored to explicit visual appearance conditions.


\section{Data}
  \label{sec:data}
  \begin{wrapfigure}{r}{0.38\linewidth}
    \centering
    \vspace{-6pt}
    \includegraphics[width=\linewidth]{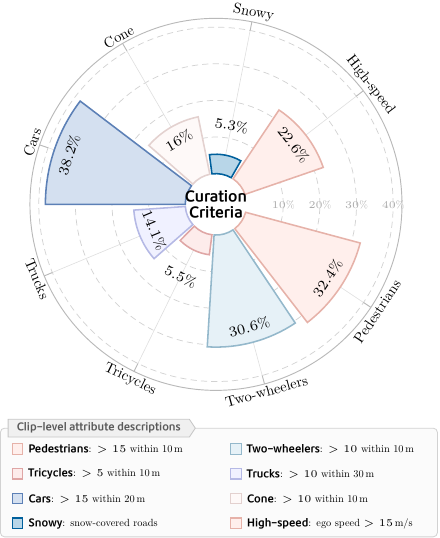}
    \caption{
      \textbf{Distribution of curated training data across clip-level attributes.}
      The polar bar chart reports the proportion of clips selected by each curation criterion, covering dense traffic participants, challenging weather, and high-speed ego motion.
    }
    \label{fig:data_balancing}
    \vspace{-40pt}
  \end{wrapfigure}
  Training a controllable driving world model requires real-world sequences that are both visually faithful and behaviorally diverse.
  We collect large-scale, high-fidelity driving sequences that cover a wide range of illumination and weather conditions, ego-vehicle behaviors, and surrounding traffic agents.
  These sequences provide the foundation for learning driving-scene dynamics and controllable responses to fine-grained conditions.
  Starting from a self-collected pool of raw driving logs, we design an automated data mining and processing pipeline to construct high-quality training data for our world model.

  \subsection{Data Curation}

    \paragraph{Sensor configuration.}
      We adopt a unified sensor configuration for all training and evaluation sequences.
      Each selected sequence is captured at 10 FPS with 10 surround-view cameras covering the full 360-degree field of view and one front-facing LiDAR sensor.
      The configuration provides synchronized visual and geometric observations for training the multi-view and multimodal world model.

    \paragraph{Data selection.}
      We further select diverse clips from the raw corpus to improve the model's ability to generate complex dynamic scenes.
      Specifically, we use detections by BEV perception models to identify scenes with dense and diverse object distributions around the ego vehicle.
      These detections provide clip-level attributes, which we use to select clips containing dense and diverse traffic participants, such as pedestrians, two-wheelers, tricycles, and large vehicles.
      In addition to the object-centric scenarios curated above, we include clips captured in various weather conditions and clips with high-speed ego-vehicle motion, covering challenging appearance and motion patterns that are necessary for driving simulation.

    \paragraph{Data balancing.}
      Finally, we balance the curated data across attributes before training.
      This balancing prevents common and simple scenarios from dominating the training distribution and preserves sufficient exposure to safety-critical and complex cases.
      The resulting attribute proportions are reported in \cref{fig:data_balancing}.
      \Cref{tab:data_summary} summarizes the final data statistics in the training and evaluation sets.

      \begin{table}[htbp]
        \centering
        \captionsetup{skip=5pt}
        \caption{
          \textbf{Dataset statistics and sensor configuration details.}
          The dataset includes both \qty{10}{\second} short clips and minute-level long clips.
          The \qty{10}{\second} training clips are used for bidirectional mid-training~(\cref{sec:bidirectional}) and teacher forcing~(TF, \cref{sec:causal-distillation}) of the initial AR causal model, while the \qty{60}{\second} long clips support iterative long-video fine-tuning~(\cref{sec:long-tuning}).
          All training and test sequences follow the same sensor configuration.
        }
        \setlength{\tabcolsep}{15pt}
        \renewcommand{\arraystretch}{1.25}
        \resizebox{.95\linewidth}{!}{%

\begin{tabular}{lccc}
  \Xhline{1.2pt}
  \multirow{2}{*}{\textbf{Statistics}} & \multicolumn{2}{c}{\textbf{Training Set}}                                                                                                                        & \multirow{2}{*}{\textbf{Test Set}}                                                                 \\
  \cline{2-3}
                                       & Mid-training \& TF                                                                                                                                               & Long-tuning                        &                                                               \\
  \Xhline{1.2pt}
  Sequence duration                    & \qty{10}{\second}                                                                                                                                                & \qty{60}{\second}                  & \qty{10}{\second}, \qty{60}{\second}                          \\
  Number of sequences                  & $\sim$\num{40000}                                                                                                                                                & $\sim$\num{4000}                   & \num{4000} (\qty{10}{\second}), \num{300} (\qty{60}{\second}) \\[.2em]
  \Xhline{.3pt}                                                                                                                                                                                                                                                                                                \\[-1.2em]
  \Xhline{.3pt}
  Sensor suite                         & \multicolumn{3}{c}{\num{10}$\times$ cameras at \qty{10}{FPS}; \num{1}$\times$ 128-beam LiDAR at \qty{10}{FPS} with a \qty{120}{\degree} FoV}                                                                                                                          \\
  Camera types                         & \multicolumn{3}{c}{\num{1}$\times$ \qty{60}{\degree} perspective, \num{5}$\times$ \qty{120}{\degree} wide-angle, and \num{4}$\times$ \qty{190}{\degree} fisheye}                                                                                                      \\
  Raw resolutions                      & \multicolumn{3}{c}{\num{576}$\times$\num{1024} images; \num{128}$\times$\num{1300} LiDAR scans}                                                                                                                                                                       \\
  \Xhline{1.2pt}
\end{tabular}

        }
        \label{tab:data_summary}
      \end{table}

  \subsection{Condition Auto-Tagging}
    The driving logs already provide several structured conditions required by our model, including sensor parameters, ego poses, and BEV maps.
    For the remaining condition inputs, we employ a fully automatic tagging pipeline that annotates each video clip with scene-level and object-level descriptions.
    \begin{itemize}
      \item \textbf{Scene-level textual prompts}: We use Qwen3-VL~\cite{baiQwen3VLTechnicalReport2025}, a vision-language model~(VLM), to generate a natural-language description for each driving video clip.
            The description focuses on global scene attributes that affect visual appearance and dynamics, including weather~(e.g., sunny, rainy, or snowy) and time of day~(e.g., daytime, dusk, or nighttime).
            Specifically, we feed the first frame of each clip to the VLM, which parses scene-level attributes from this reference image.
      \item \textbf{Object-level descriptions}: The BEV model provides the 3D bounding box and semantic category for each object.
            For each nearby object, we crop the object region from the video using its labeled 3D bounding box and use the resulting crop as the corresponding image description.
            We then prompt the VLM to annotate each object with a natural-language description, focusing on appearance, size, and state.
    \end{itemize}

    Overall, the resulting dataset provides a holistic foundation for training long-horizon driving world models.
    It contains synchronized multi-view videos and aligned LiDAR observations, enabling joint camera--LiDAR generation under consistent sensor geometry.
    The clips span both 10-second segments for dense short-term dynamics and minute-level sequences for long-horizon rollout.
    Together with the additional object-level annotations, the dataset supports fine-grained control over individual objects beyond geometric placement.


\section{Controllability Evaluation Using a VLM Judge}
  \label{sec:metric}

  \begin{wrapfigure}{r}{0.44\linewidth}
    \centering
    \vspace{-10pt}
    \includegraphics[width=\linewidth]{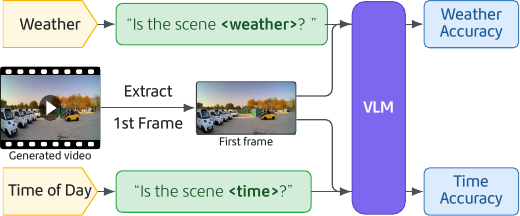}
    \caption{Scene controllability evaluation.}
    \label{fig:vlm_judge_scene}
    \vspace{-10pt}
  \end{wrapfigure}

  We evaluate the generated driving videos from two complementary perspectives.
  First, we adopt standard distribution-level metrics, including FID~\cite{heuselGANsTrainedTwo2018} and FVD~\cite{unterthinerAccurateGenerativeModels2019}, to measure the perceptual realism of the full video rollout.
  These metrics reflect the fundamental video generation capability of the model, but do not directly measure whether a driving world model follows the rich control signals that define the target simulation.
  To evaluate such fine-grained controllability, we introduce an automated VLM-based judge that leverages the visual and semantic knowledge encoded in foundation models to examine condition adherence in generated driving scenes.
  Our evaluation pipeline covers three dimensions: scene-level controllability, view-wise object-level controllability, and cross-view object-level consistency.

  \subsection{Scene Controllability}
    \label{sec:scene-level-controllability}

    We first evaluate whether the generated sequence follows the requested global scene attributes, including weather and time of day.
    For each generated rollout, we query the VLM with binary questions derived from the target scene prompt.
    For example, under a \emph{sunny daytime} condition, we ask two questions: ``Is this scene \emph{sunny}?
    Answer yes or no.
    '' and ``Is this scene \emph{daytime}?
    Answer yes or no.
    ''
    The two yes-or-no responses are used to measure weather correctness and time-of-day correctness, respectively.

  \subsection{View-Wise Object Controllability}
    \label{sec:monocular-object-level-controllability}

    \begin{wrapfigure}{r}{0.44\linewidth}
      \centering
      \vspace{-10pt}
      \includegraphics[width=\linewidth]{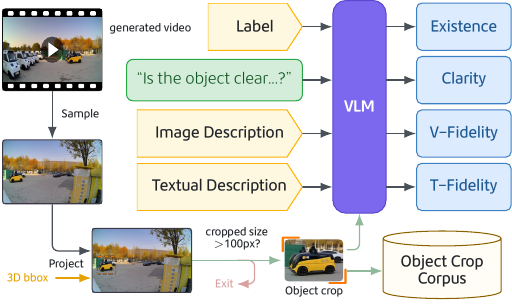}
      \caption{View-wise object controllability evaluation.}
      \label{fig:vlm_judge_monocular}
      \vspace{-15pt}
    \end{wrapfigure}

    Beyond global scene attributes, our model aims to support fine-grained control over each traffic object around the ego vehicle.
    Each object is specified through four complementary conditioning modalities: its 3D bounding box, a semantic category, an image description, and a textual description.
    During evaluation, we take a video generated for one camera view and the four conditioning inputs associated with a given object track ID, and then uniformly sample frames from the video.
    For each sampled frame, we project the conditioned 3D bounding box onto the image plane and crop the corresponding region.
    We keep only valid crops whose shorter side is larger than \num{100} pixels, and evaluate each crop using the following view-wise object-level prompts:

    \begin{itemize}
      \item \textbf{Existence.}
            We prompt the VLM to verify whether the cropped region contains an object of the specified semantic category: ``Is this object a \texttt{[object\_label]}?
            Answer yes or no.
            ''
      \item \textbf{Clarity.}
            We ask the VLM to judge whether the object in the crop is visually recognizable: ``Is this object clearly discernible?
            Answer yes or no.
            ''
      \item \textbf{Visual fidelity.}
            When the object is conditioned on an image crop, we use the crop as the reference appearance and ask the VLM to compare it with the generated crop: ``Are these two images of the same object?
            Answer yes or no.
            ''
      \item \textbf{Textual fidelity.}
            When a textual description is provided for the object, we ask the VLM to determine whether the generated crop matches the description: ``Does this image match the description \texttt{[textual\_description]}?
            Answer yes or no.
            ''
    \end{itemize}

  \subsection{Cross-View Object Consistency}
    \label{sec:cross-view-object-level-controllability}

    \begin{wrapfigure}{r}{0.44\linewidth}
      \centering
      \vspace{-10pt}
      \includegraphics[width=\linewidth]{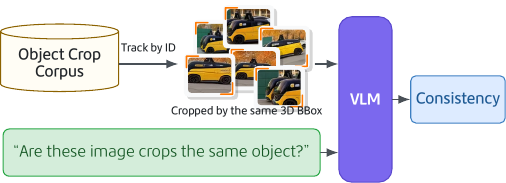}
      \caption{Cross-view object consistency evaluation.}
      \label{fig:vlm_judge_cross_view}
      \vspace{-15pt}
    \end{wrapfigure}

    The same traffic object should remain visually consistent across time and viewpoints under the shared 3D object condition.
    During evaluation, we uniformly sample frames from the generated videos of multiple views.
    For each timestamp, we project the conditioned 3D bounding box associated with the same object track ID onto all camera views and crop the corresponding object regions.
    We keep valid crops whose shorter side is larger than \num{100} pixels, and feed all valid crops from the same timestamp and track ID into the VLM jointly.
    We prompt the VLM with the question, ``Are these image crops of the same object?
    Answer yes or no.
    ''
    The binary response measures whether the generated multi-view observations preserve a consistent object identity across viewpoints.

    \paragraph{Scoring.}
      For each prompt type, we report the frequency with which the VLM answers ``yes'' as the corresponding controllability score.
      A higher score indicates that the generated rollout more reliably satisfies the evaluated condition.
      We additionally aggregate object-level scores by semantic category, which provides a category-wise assessment of the model's ability to generate and control different types of traffic objects.


\section{Experiments and Results}

  \subsection{Basic Evaluation}
    \label{sec:basic-evaluation}
    \paragraph{Multi-view and multimodal generation.}

      \begin{wraptable}{r}{0.36\linewidth}
        \centering
        \vspace{-10pt}
        \captionsetup{skip=5pt}
        \caption{
          \textbf{Quantitative evaluation of basic generation quality.}
        }
        \setlength{\tabcolsep}{8pt}
        \renewcommand{\arraystretch}{1.25}
        \resizebox{\linewidth}{!}{%

\begin{tabular}
  {lcc}
  \Xhline{1.2pt}
  \textbf{Method}                                           & \textbf{FID} $\downarrow$ & \textbf{FVD} $\downarrow$ \\
  \Xhline{1.2pt}
  MagicDriveV2~\cite{gaoMagicDriveV2HighResolutionLong2024} & \tablenum{41.7}           & \tablenum{346.1}          \\
  \makecell[l]{\textbf{Ours}}
                                                            & \tablenum{34.8}           & \tablenum{288.7}          \\
  \Xhline{1.2pt}
\end{tabular}

        }
        \label{tab:quantitative-basic-quality}
        \vspace{-5pt}
      \end{wraptable}

      We first evaluate the bidirectional driving-world-model teacher trained in \cref{sec:bidirectional}.
      We use generated \qty{10}{\second} driving videos to assess its generation quality and fine-grained controllability.

\begin{figure}[htbp]
        \centering
        \vspace{25pt}
        \includegraphics{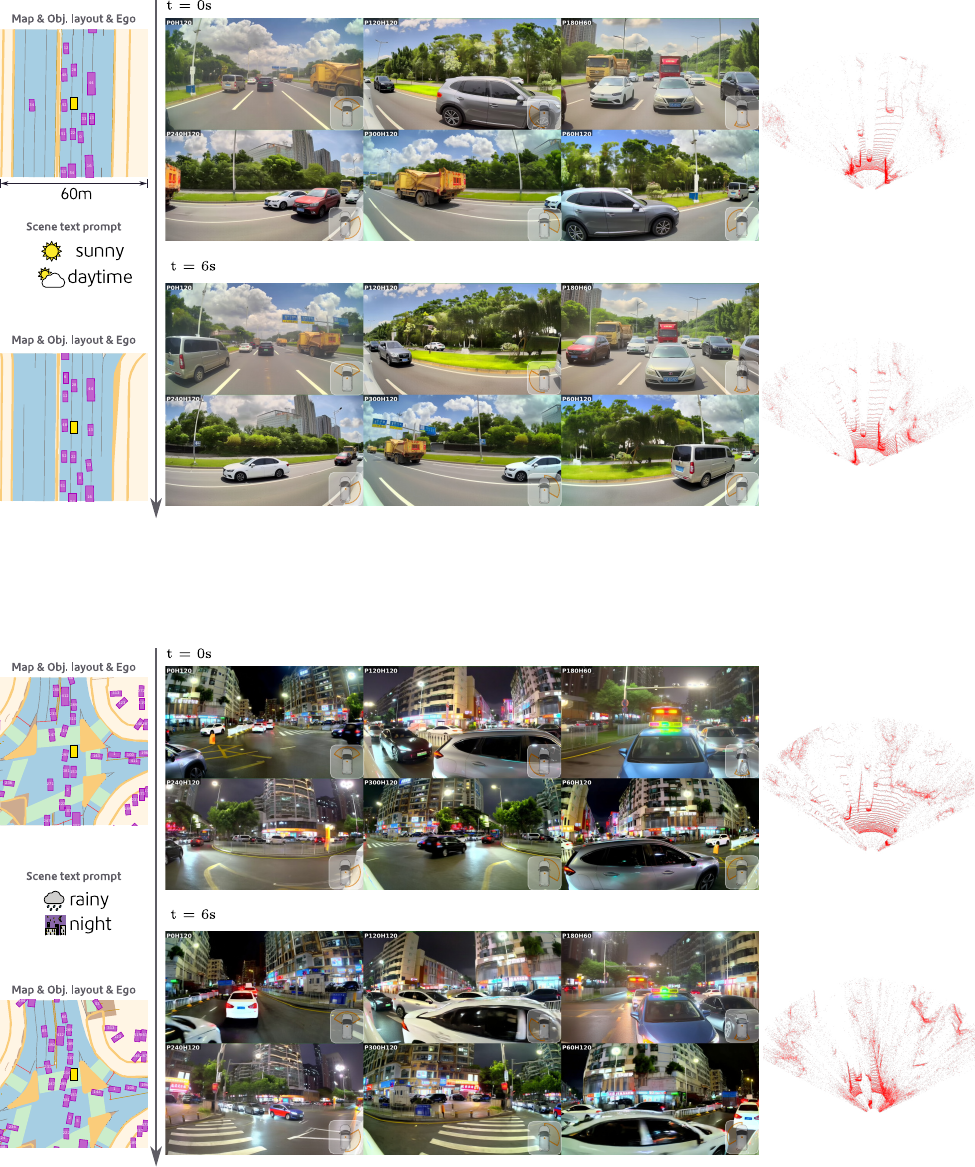}
        \caption{Qualitative results of multi-view and multimodal generation.
          From left to right, we show the input control signals, generated multi-camera streams, and the LiDAR scan at the synchronized timestamp.
        }
        \label{fig:multi-modal}
      \end{figure}

      \Cref{fig:multi-modal} presents two generated driving clips, each visualized using six of the ten camera views and one LiDAR scan, demonstrating coherent cross-view and cross-modal synthesis.
      For a quantitative comparison of generation quality, we report FID~\cite{heuselGANsTrainedTwo2018} and FVD~\cite{unterthinerAccurateGenerativeModels2019} against MagicDriveV2~\cite{gaoMagicDriveV2HighResolutionLong2024}; both models are trained on the same dataset.
      As shown in \cref{tab:quantitative-basic-quality}, our method achieves lower FID and FVD through a unified design: self-attention performs long-range spatiotemporal modeling, and shared cross-attention injects the encoded control signals.
      These improvements also suggest that incorporating fine-grained object-level conditioning effectively enhances visual quality.

    \paragraph{Fine-grained controllability.}

\begin{figure}[htbp]
        \centering
        \vspace{25pt}
        \includegraphics{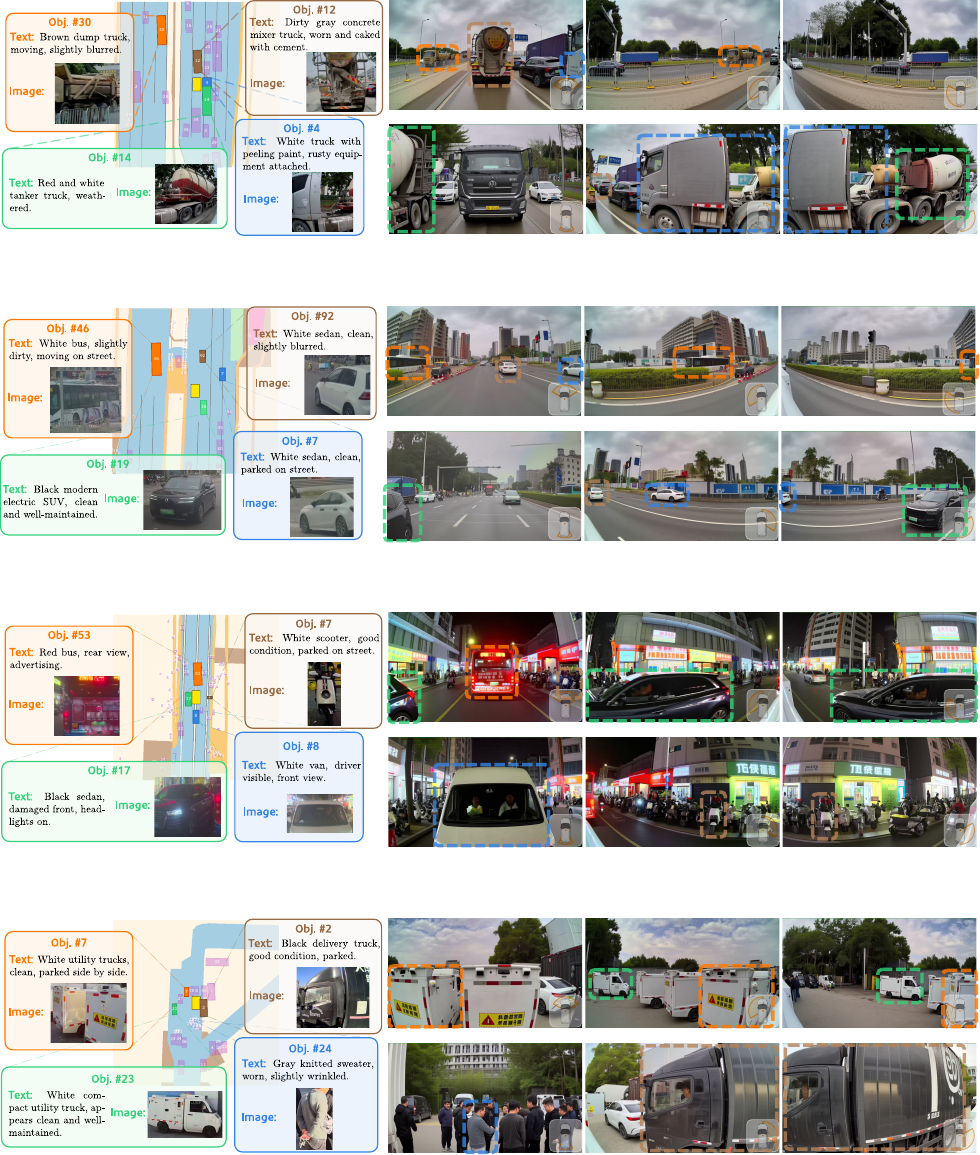}
        \caption{Qualitative evaluation of fine-grained object control.
          For each case, the left panel shows the input textual and visual descriptions for four objects near the ego vehicle.
          In the six-view generation results on the right, color-coded bounding boxes highlight the corresponding conditioned objects.
        }
        \label{fig:fine-grained-controllability}
      \end{figure}

      In \cref{fig:fine-grained-controllability}, we qualitatively visualize the fine-grained object controllability of \method{}.
      While preserving object positions and orientations specified by the conditions, our method also controls appearance-level attributes such as shape and local visual details.
      We then assess the condition-following performance using the metrics introduced in \cref{sec:metric}.
      As a baseline, we use MagicDriveV2~\cite{gaoMagicDriveV2HighResolutionLong2024} with inputs that exclude object-level textual and image descriptions.
      As shown in \cref{tab:quantitative-fined-grained-control}, \method{} not only follows object-level textual and visual conditions effectively, but also benefits from these fine-grained conditions to further improve scene control, object-level spatial controllability, visual sharpness, and cross-view consistency.

      \begin{table}[htbp]
        \centering
        \captionsetup{skip=5pt}
        \caption{
          \textbf{Quantitative evaluation of fine-grained controllability.}
          Our model achieves superior controllability over MagicDriveV2~\cite{gaoMagicDriveV2HighResolutionLong2024} in both scene-level and object-level control, while maintaining high cross-view object consistency.
          We report \emph{Oracle} results by evaluating the ground-truth test videos, which provide empirical upper bounds for the controllability metrics.
        }
        \setlength{\tabcolsep}{5pt}
        \renewcommand{\arraystretch}{1.15}
        \resizebox{\linewidth}{!}{%

\begin{tabular}
  {lccccccc}
  \specialrule{1.4pt}{0pt}{2pt}
  \multirow{2}{*}{\textbf{Method}} & \multicolumn{2}{c}{\textbf{Scene Controllability} (\%) $\uparrow$} & \multicolumn{4}{c}{\textbf{View-Wise Object Evaluation} (\%) $\uparrow$} & \textbf{Cross-View Evaluation}                                                                                                                         \\
  \cmidrule(lr){2-3}\cmidrule(lr){4-7}\cmidrule(lr){8-8}
                                   & \textbf{Weather}                                                   & \textbf{Time of Day}                                                     & \textbf{Existence}             & \textbf{Clarity} & \textbf{Visual Fidelity} & \textbf{Textual Fidelity} & \textbf{Object Consistency} (\%) $\uparrow$ \\
  \Xhline{1.2pt}
  \rowcolor{gray!10}
  \emph{Oracle}                    & \tablenum{76.7}                                                    & \tablenum{83.0}                                                          & \tablenum{95.4}                & \tablenum{74.6}  & \tablenum{87.0}          & \tablenum{75.4}           & \tablenum{93.3}                             \\
  \makecell[l]{MagicDriveV2~\cite{gaoMagicDriveV2HighResolutionLong2024}}
                                   & \tablenum{75.4}                                                    & \tablenum{79.1}                                                          & \tablenum{82.6}                & \tablenum{59.1}  & \tablenum{13.4}          & \tablenum{11.6}           & \tablenum{78.9}                             \\
  \makecell[l]{\textbf{Ours}}
                                   & \tablenum{76.6}                                                    & \tablenum{80.9}                                                          & \tablenum{90.6}                & \tablenum{67.2}  & \tablenum{62.7}          & \tablenum{59.1}           & \tablenum{84.5}                             \\
  \Xhline{1.2pt}
\end{tabular}

        }
        \label{tab:quantitative-fined-grained-control}
      \end{table}

    \paragraph{Long-horizon streaming.}

      \begin{figure}
        \centering
        \includegraphics{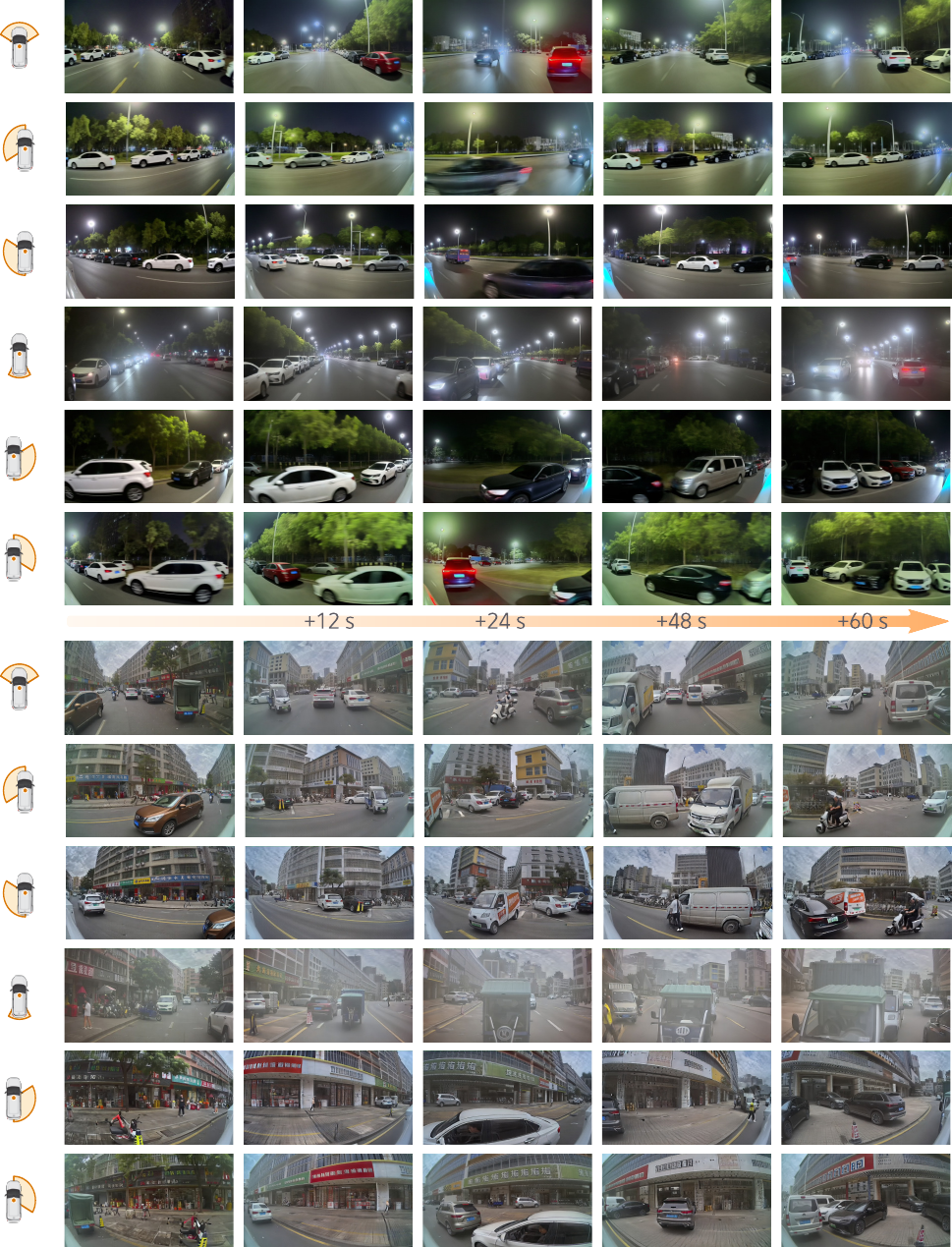}
        \caption{Qualitative evaluation of long-horizon rollouts.
          We present key frames from 60-second driving videos generated in parallel across six views.
          The rollouts remain stable over long horizons, preserving coherent motion and visual appearance without catastrophic drift.
        }
        \label{fig:long-streaming}
      \end{figure}

      \begin{figure}[htbp]
        \centering
        \includegraphics{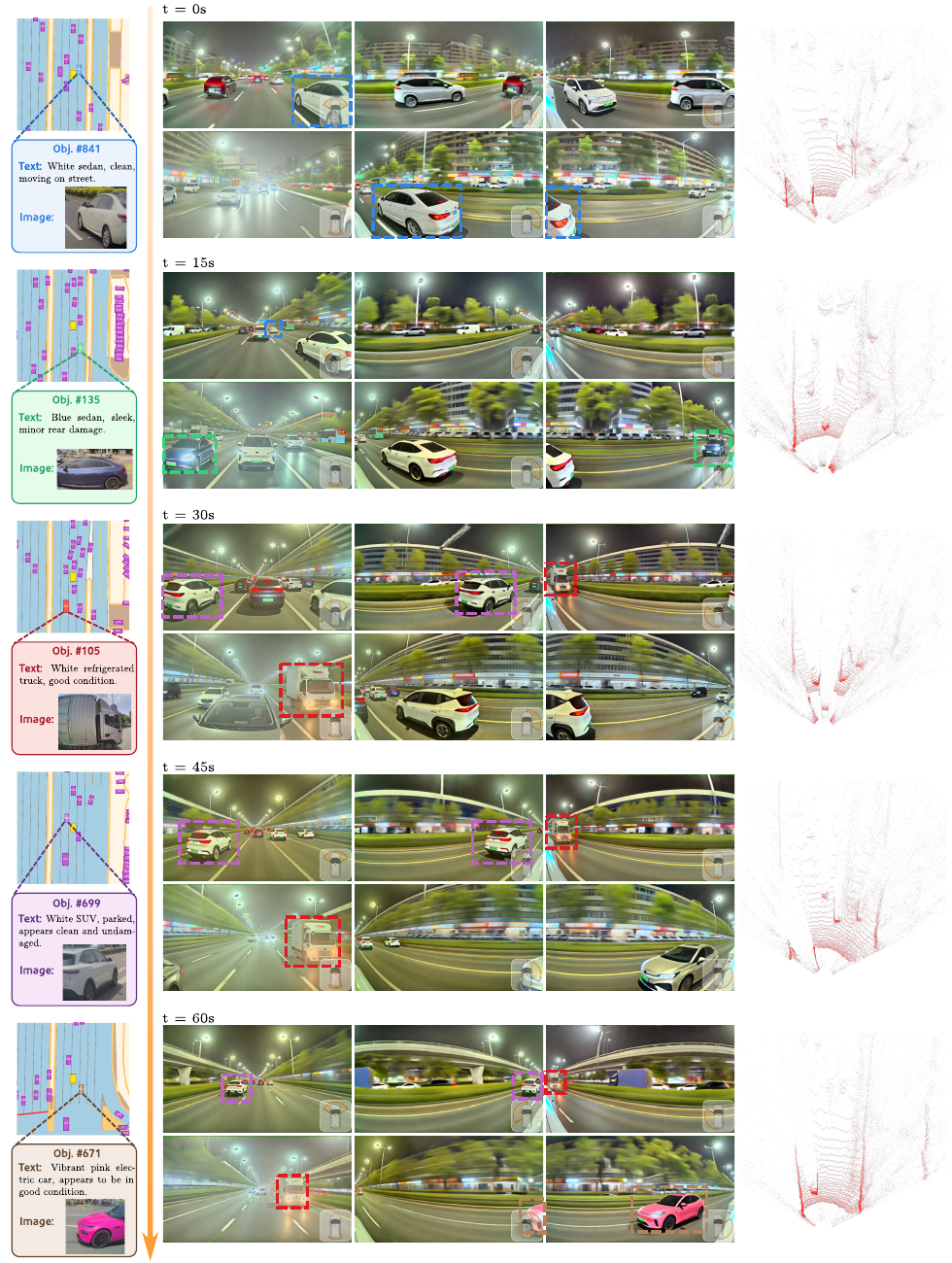}
        \caption{Holistic qualitative results demonstrating minute-long rollouts with fine-grained object controllability.}
        \label{fig:holistic-showcase}
      \end{figure}

      \begin{figure}
        \centering
        \includegraphics{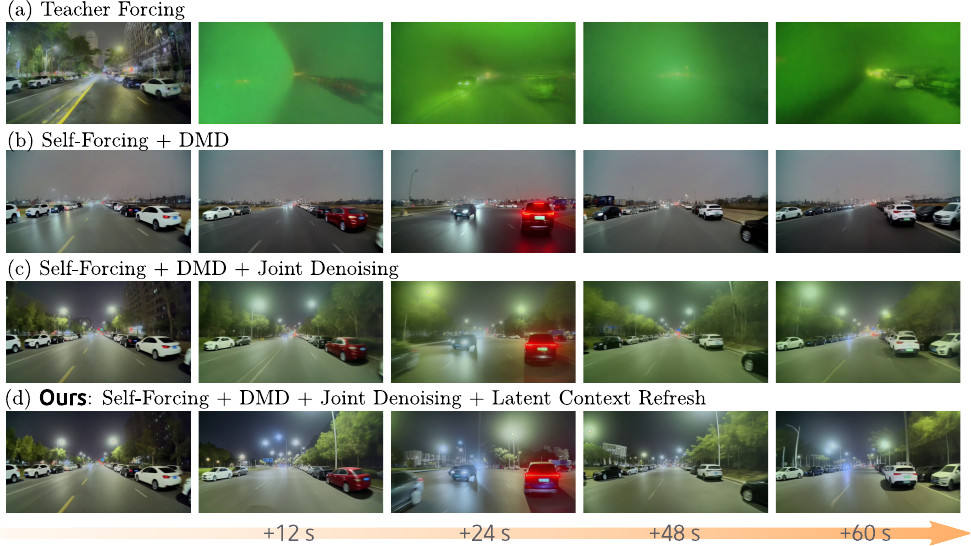}
        \caption{
          Visual ablation of key components in our long-horizon generation training recipe.
          Our full model maintains clearer scene structure and more stable visual quality throughout autoregressive rollout.
        }
        \label{fig:long-streaming-ablation}
      \end{figure}

      As shown in \cref{fig:long-streaming}, \method{} generates two \qty{60}{\second} multi-camera rollouts at $1024\times567$ and 10 FPS, while maintaining coherent motion and appearance over long horizons without catastrophic drift.
      \Cref{fig:holistic-showcase} presents holistic results on minute-long rollouts exhibiting fine-grained object controllability.
      In terms of throughput, our four-step student denoiser achieves 0.7, 2.3, and 7 FPS at $1024\times567$, $424\times800$, and $224\times400$, respectively, for six camera views plus one synchronized LiDAR stream on eight A100 GPUs.
      Furthermore, \cref{fig:long-streaming-ablation} visualizes the impact of key components in our long-horizon generation training recipe.
      The results indicate that supplementing the DMD objective with the supervised denoising loss $\mathcal{L}_{\mathrm{denoise}}$ on ground-truth video latents mitigates background degradation, whereas latent context refresh reduces inter-chunk flickering artifacts.

  \subsection{Case Study: A Long-Tail Driving Scenario}

    One key value of autonomous-driving simulation lies in controllably enriching the diversity of training data for long-tail scenarios.
    Although the over thousands of hours of driving data collected each day can continually improve perception foundation models on common cases, these models still struggle to generalize to rare scenarios that are underrepresented in the captured data.
    In practical workflows, such long-tail cases are usually defined case by case with concrete attributes: the scenario itself is rare, the failure mode to be addressed is specific, and acquiring sufficient real-world data is prohibitively expensive.

\begin{figure}[htbp]
      \centering
      \includegraphics{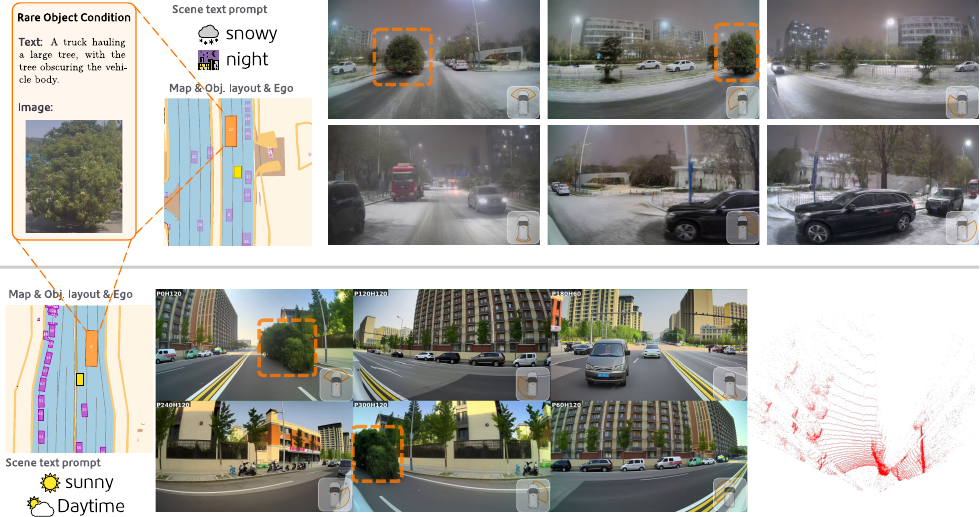}
      \caption{Qualitative evaluation on long-tail scenario augmentation.
        We insert the same long-tail object—a tree-hauling truck—into two different scene configurations, thereby increasing the diversity of the target scenario.
      }
      \label{fig:a-long-tail-case-study}
    \end{figure}

    We study a representative case of a truck hauling large trees.
    From the rear view, the truck body is almost completely occluded by tree branches, causing its appearance to differ substantially from that of ordinary vehicles and potentially posing a severe challenge to downstream perception.
    However, this case is extremely scarce in real logs, with fewer than five clips observed in hundreds of thousands of hours of data.
    While our world model is trained on regular driving data, its fine-grained object-control interface, together with the efficient post-training procedure introduced in \cref{sec:longtail}, allows the model to adapt from only a few rare clips.
    \Cref{fig:a-long-tail-case-study} shows two variants of the tree-hauling truck synthesized by the resulting driving simulator.
    The results demonstrate that providing textual and image descriptions of the truck to the few-clip-adapted simulator is sufficient to generate the target long-tail scenario under diverse conditions.

    Furthermore, we synthesize \num{500} \qty{10}{\second} video clips of tree-hauling trucks and mix them with \num{50000} real clips to construct an augmented training set.
    We then train a BEV detection model on this dataset and evaluate it on a test set containing 99 annotated boxes of tree-hauling trucks.
    \Cref{tab:long-tail-bev-detection} shows that these synthetic clips serve as effective data augmentation, substantially improving perception performance on this specific long-tail case while maintaining comparable regular-set detection performance.

    \begin{table}[htbp]
      \centering
      \captionsetup{skip=5pt}
      \caption{
        \textbf{BEV detection results on the tree-hauling truck long-tail case.}
        Synthetic clips generated by the few-clip-adapted simulator improve recall on the target long-tail object while preserving regular-set detection performance.
      }
      \setlength{\tabcolsep}{14pt}
      \renewcommand{\arraystretch}{1.15}
      \resizebox{.8\linewidth}{!}{%

\begin{tabular}
  {lcc}
  \Xhline{1.2pt}
  \textbf{BEV Detector Training Data}  & \textbf{Tree-Hauling Truck Recall} $\uparrow$ & \textbf{Regular-Set mAP} $\uparrow$ \\
  \Xhline{1.2pt}
  50k real clips                       & 1.0\%                                         & 66.7\%                              \\
  50k real clips + 500 synthetic clips & 69.7\%                                        & 66.8\%                              \\
  \Xhline{1.2pt}
\end{tabular}

      }
      \label{tab:long-tail-bev-detection}
    \end{table}

    \begin{figure}[htbp]
      \centering
      \includegraphics{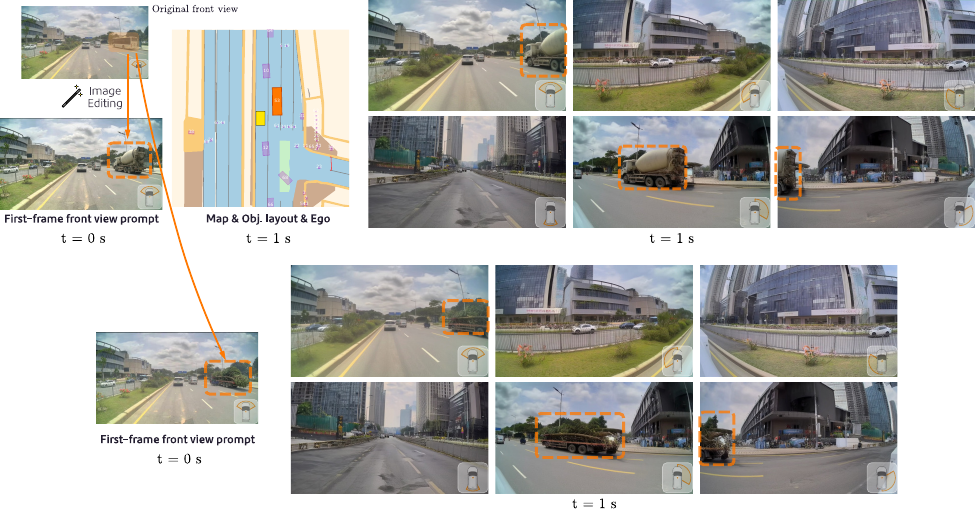}
      \caption{Qualitative results of vision-conditioned generation prompted with an edited first-frame front view.
        Left: the original front view at $t{=}0$ and the edited prompt image after object replacement.
        Right: generated results at $t{=}1\,\mathrm{s}$ conditioned on the edited prompt.
        The top and bottom rows show the original bus replaced by a cement mixer truck and a tree-hauling truck, respectively.
      }
      \label{fig:first-frame-condition}
    \end{figure}

  \subsection{First-Frame-Conditioned Generation}

    Finally, we demonstrate customizable driving simulation through first-frame editing using the fine-tuned visual reference-conditioned model introduced in \cref{sec:visual_conditioned_suite}.
    Given an original front-view image, we use image editing tools to replace any object with an uncommon target object, such as a cement mixer truck or a tree-hauling truck, as illustrated in \cref{fig:first-frame-condition}.
    Conditioned on this edited first-frame front view, the model completes the remaining camera views and predicts future frames according to the subsequent time-evolving control signals.
    In these examples, the replacement target is specified through single-view editing, and its subsequent evolution remains controllable through dense temporal signals.
    This process provides a zero-shot, controllable pathway for generating long-tail driving data.


\section{Conclusion}

  This paper has introduced \method{}, a generative driving world model suitable for autonomous-driving simulation.
  One central goal of \method{} is to move beyond geometry-only scene specification: the model generates synchronized multi-view camera streams and LiDAR range maps while conditioning individual traffic agents on 3D layout, category, visual appearance, and textual attributes.
  This object-centric interface is implemented within a shared DiT latent backbone and made practical for streaming rollout through a progressive training pipeline.
  For streaming generation, the developed training recipe transfers a bidirectional video prior to a low-latency causal student, exposes the student to self-generated histories, and improves inter-chunk consistency through latent context refresh.
  On top of our base model, efficient few-clip post-training and visual reference-conditioned variants provide two complementary paths for constructing customized long-tail scenarios.

  Across generation quality, condition following, rollout stability, and downstream augmentation, the results support the effectiveness of controlling objects at both the spatial-layout and visual-appearance levels for driving simulation.
  Compared with an existing controllable driving-video generation baseline, \method{} improves FID/FVD from 41.7/346.1 to 34.8/288.7, increases object visual and textual fidelity from 13.4\% and 11.6\% to 62.7\% and 59.1\%, respectively, and raises cross-view object consistency from 78.9\% to 84.5\%.
  The distilled few-step causal student can generate stable 60-second rollouts, demonstrating that fine-grained conditioning can be retained in long autoregressive generation.
  We also demonstrate that controllable long-tail synthesis enabled by efficient few-clip post-training can yield measurable perception gains: in a representative tree-hauling-truck case, adding \num{500} synthetic clips to \num{50000} real clips improves target recall from 1.0\% to 69.7\% while preserving regular-set mAP\@.
  More broadly, \method{} suggests a path toward simulation systems in which rare events are specified not only by where objects appear, but also by what they look like and how they persist across views, modalities, and time.
  Future work should expand the coverage of rare-event taxonomies and evaluate the model in genuine closed-loop settings, where learned driving policies interact with the generated environment and are assessed for their responses to safety-critical events.


\bibliographystyle{author-kit/ieeenat_fullname}
\bibliography{src/main/reference}

\end{document}